\documentclass{article}

    \usepackage[preprint]{neurips_2021}

\usepackage{float}
\restylefloat{table}
\usepackage[utf8]{inputenc} 
\usepackage[T1]{fontenc}    
\usepackage{hyperref}       
\usepackage{url}            
\usepackage{booktabs}       
\usepackage{amsfonts}       
\usepackage{nicefrac}       
\usepackage{microtype}      
\usepackage{xcolor}
\usepackage{graphics}
\usepackage{graphicx}
\usepackage{physics}
\title{Video2Skill: Adapting Events in Demonstration Videos to Skills in an Environment using Cyclic MDP Homomorphisms}

\author{%
  Sumedh A. Sontakke\thanks{Corresponding Author: ssontakk@usc.edu} \\
  University of Southern California\\
   \And
   Sumegh Roychowdhury \\
   Indian Institute of Technology, Kharagpur \\
   \AND
   Mausoom Sarkar \\
   Adobe MDSR \\
   \And
   Nikaash Puri\\
   Adobe MDSR\\
   \And
   Laurent Itti\\
   University of Southern California\\
   \And
   Balaji Krishnamurthy\\
   Adobe MDSR\\
}
\def\approachName{V2S\space}
\newtheorem{definition}{Definition}
\begin{document}

\maketitle

\begin{abstract}
Humans excel at learning long-horizon tasks from demonstrations augmented with textual commentary, as evidenced by the burgeoning popularity of tutorial videos online. Intuitively, this capability can be separated into 2 distinct subtasks - first, dividing a long-horizon demonstration sequence into semantically meaningful events; second, adapting such events into meaningful behaviors in one's own environment. Here, we present Video2Skill (V2S), which attempts to extend this capability to artificial agents by allowing a robot arm to learn from human cooking videos. We first use sequence-to-sequence Auto-Encoder style architectures to learn a temporal latent space for events in long-horizon demonstrations. We then transfer these representations to the robotic target domain, using a small amount of offline and unrelated interaction data (sequences of state-action pairs of the robot arm controlled by an expert) to adapt these events into actionable representations, i.e., skills. Through experiments, we demonstrate that our approach results in self-supervised analogy learning, where the agent learns to draw analogies between motions in human demonstration data and behaviors in the robotic environment. We also demonstrate the efficacy of our approach on model learning - demonstrating how Video2Skill utilizes prior knowledge from human demonstration to outperform traditional model learning of long-horizon dynamics. Finally, we demonstrate the utility of our approach for non tabula rasa decision making, i.e, utilizing video demonstration for zero-shot skill generation.               
\end{abstract}

\section{Introduction and Related Work}

Offline reinforcement learning has been of substantial interest to the community with continued efforts in attempting to teach RL agents to perform tasks simply from a corpus of expert demonstration data (\cite{levine2020offline,kidambi2020morel,agarwal2020optimistic}). While offline RL holds promise, it is challenging because it requires making counter-factual queries under distributional shift (i.e., the agent cannot explore the effects of hypothetical action sequences not present in the training data; \cite{levine2020offline}). Additionally, current offline RL formulations make 2 constraining assumptions - first, they require \textbf{domain coincidence}, i.e., that the state and action spaces of the demonstrations and the downstream agent being trained coincide. This can be restrictive especially in applications to domains such as robotics, where expert demonstrations in the same domain may not be available. Consider, for example, attempting to teach a robot to perform medical surgery. In such a scenario, current offline RL formulations would fail as they would require a dataset of demonstrations from an expert robot performing surgery. However, generating such an expert agent using vanilla RL in a safety-critical application would be disastrous due to exploration - a classic chicken-or-egg conundrum. Instead, it is more likely that   demonstrations from a human expert would be available, and thus we need new methods to become able to exploit them.   

Second, offline RL assumes \textbf{task coincidence}, i.e., it attempts to train agents to perform the same tasks as made available in the demonstration dataset, e.g., a demonstration of a robotic manipulation task, say grasping, will result in a policy that enables an agent to grasp. 
Combined, these assumptions mean that offline RL assumes that the MDPs on which the expert demonstrates its behavior and on which the downstream agent are trained to behave in are the same. 

Hence, these assumptions make offline RL difficult to apply to practical robotic scenarios. In this work, we attempt to relax these assumptions. We  utilize the large corpus of online human video tutorials of complex long horizon tasks to teach a robotic agent to perform semantically meaningful behaviors in its own environment. Inspired by \cite{zhu2017unpaired}, our work attempts to learn adaptible short-horizon motion representations - called events -  using domain randomization on human demonstrations. We then utilize a small amount of environment-specific demonstration data to adapt this latent space for domain specific behavior.

We improve upon the state-of-the-art in the following ways:

\textbf{Unsupervised Event Representation Learning:} Learning temporal representations from demonstrations (\cite{chen2019towards}, \cite{boggust2019grounding}, \cite{tosi2020distilled}) (e.g., skill learning, event detection, etc.) typically requires large datasets of demonstrations, with expensive human annotations for timestamps corresponding to each event. \approachName, on the contrary, learns event representations without temporal supervision, i.e., it divides long-horizon trajectories into semantically meaningful subsequences, without access to any temporal annotations that splits these trajectories.

\textbf{Domain and Task Invariance:} \approachName abstracts events from demonstrations of a variety of cooking tasks. Additionally, these videos originate from a number of sources, varying in camera angles, instructional styles, etc. Thus, through domain randomization, our architectures generate domain invariant event representation. Unsupervised skill learning typically has more restrictive assumptions (\cite{Shankar2020Discovering},\cite{eysenbach2018diversity},\cite{sharma2019dynamics}) requiring that demonstration data originate from a single domain, with the same state and action spaces. 

\textbf{Offline and Reward-free Skill Learning:} Unsupervised skill discovery (\cite{eysenbach2018diversity,sharma2019dynamics,xu2018neural}), \cite{huang2019neural}) also typically requires costly interactions with an environment to discover skill sequences. Such assumptions can be infeasible in domains such as healthcare, where active exploration may not only be impossible, but potentially dangerous. \cite{eysenbach2018diversity} learn a large number of low-level sequences of actions by enforcing that the corpus of skills acquired is diverse. Similarly, \cite{sharma2019dynamics} attempt to learn skills such that under a skill, subsequent transitions are almost deterministic in a given environment. \approachName first discovers event representations from freely available human demonstration data, and subsequently adapts them to learn environment-specific skills. 

\textbf{Long Horizon Learning from Demonstration:} Long-horizon tasks remain the bane of decision-making algorithms, especially in the offline-learning scheme, due to an aggregation of sub-optimal behaviors over a horizon (\cite{nicolescu2003natural}). Imitation learning (\cite{esmaili1995behavioural},  \cite{atkeson1997robot}, \cite{schaal1997learning}, \cite{pastor2009learning}, \cite{peters2013towards}, \cite{niekum2012learning}) has shown how agents can learn simple tasks from demonstrations. More recently, \cite{schmeckpeper2019learning} shows that agents can learn to maximize external reward using a large corpus of observation data, i.e., trajectories of states, and a relatively smaller corpus of interaction data, i.e., trajectories of state-action pairs. However, such approaches are restricted to short horizons, while \approachName is able to generate skills for long-horizon tasks like cooking.   

\textbf{Multi-modal World Models:} \approachName learns representations for events occurring free-flowing tutorial videos utilizing both textual and visual inputs which are available in typical human demonstrations. Once adapted to a specific environment, it describes the model of the environment. We show that these World Models outperform typical model (\cite{higuera2018synthesizing,nagabandi2018neural, lakshminarayanan2016simple, chua2018deep}) learning methods on multi-step prediction. 

\textbf{Incorporating Prior Knowledge into Decision Making:} We propose a adaptation based method to incorporate prior knowledge into decision making - both for model-based and model-free RL. We pre-train a Backbone network on real world cooking data and subsequently learn environment-specific adapter functions to model dynamics of a kitchen environment. We show how the pre-training aids efficient dynamics learning and yields semantically meaningful representations.
\section{Methods}
\label{sec:Methods}
Tutorial videos contain informative demonstrations of complex real-world tasks. These consist of humans acting as expert agents in an abstract Markov Decision Process (MDP). While we do not have access to the state and action spaces of such an abstract MDP, we do, however, have access to proxies for them through the video frames and textual commentary in the tutorial video, i.e., \textbf{state}$\rightarrow$ \textbf{action}~ \text{is like}~\textbf{video frame}$\rightarrow$ \textbf{commentary}.   

These video demonstrations consist of several events which are described in words and viewed through short sequences of video frames.
We utilize such real world human demonstration data to learn environment agnostic event representations. We do this using domain randomization - by training a multi-modal temporal auto-encoder-style architecture (called \textbf{Backbone network}) on human cooking demonstrations consisting of a variety of cooking recipes and tasks. Additionally, our data comes from many sources - with a variety of camera angles, lighting, etc. The temporal autoencoder thus generates domain and task independent embeddings for sequences of videos and words. Thus, given a human demonstration of say, poaching eggs, our architecture can isolate semantically meaningful subsequences, like cracking an egg, pouring water, etc. These event representations encode both a sequence of observations in the domain of the human cooking videos, and the associated "action sequences" in the form of textual tutorial commentary. 

We then utilize a small amount of demonstration data in the environment of a real robotic agent. This data consists of sequences of states and actions of an expert robot demonstrating related tasks in the environment; for example, the robot arm opening cabinet doors, etc., but not cooking. We then force the representations of these robotic demonstrations to be in same space as those of human cooking demonstrations. This is achieved using a pair of MDP Homomorphisms which map the robotic MDP to the human abstract MDP and vice-versa. The homomorphisms are learnt in a cyclical manner, thereby requiring no supervision during training. Using the homomorphisms, we can later translate cooking events (e.g., cracking an egg) into the target robotic space and \textit{vice-versa}, resulting in zero-shot skill generation. Thus, \textbf{state}$\rightarrow$ \textbf{action}$\rightarrow$ \textbf{skill}~\text{is analogous to} \textbf{ video-frame}$\rightarrow$ \textbf{commentary} $\rightarrow$ \textbf{event}.

\subsection{Event Representation Learning from Demonstration Videos}
\begin{figure}[t]
    \centering
    \includegraphics[width = \textwidth, height=110pt]{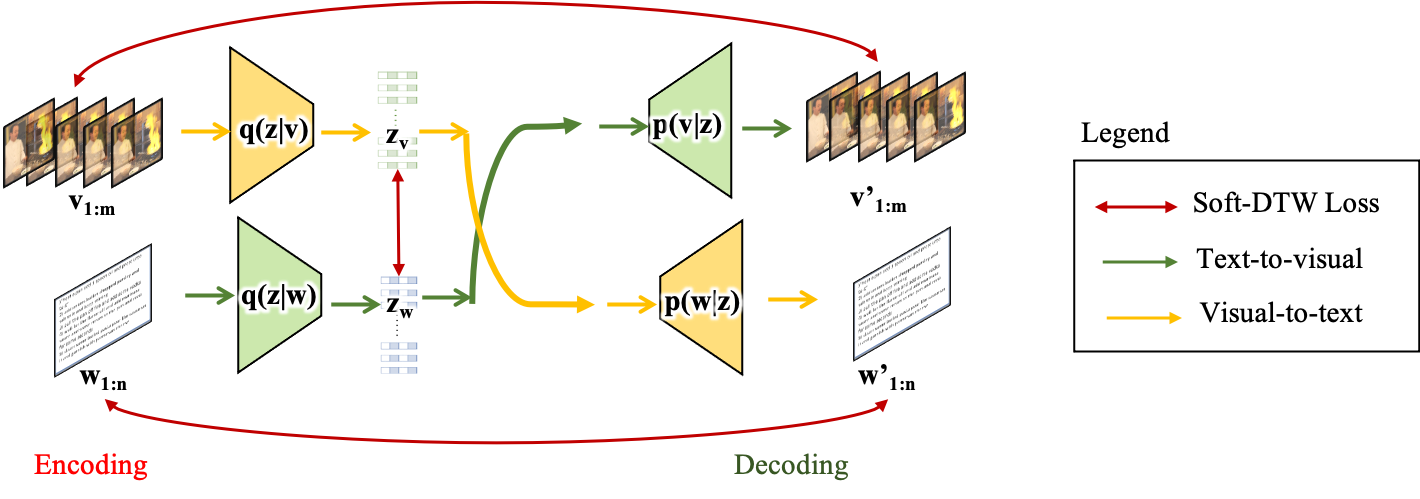}
    \caption{\textbf{Backbone Network.} \approachName contains a backbone temporal autoencoder which learns a semantically-meaningful embedding space, encoding events that occur in natural free-flowing tutorial videos, without explicit temporal supervision for event start and end timestamps. Section~\ref{sec:Methods} contains details of components.}
    \label{fig:backbone}
\end{figure}

Intuitively, we define an event as a short sequence of states which may occur repeatedly across several demonstration trajectories.
Events have an upper limit on their length in time steps.
They can be obtained from both a sequence of demonstration images (video data) ($\vb{V} = \vb{v}_{0:m}$) and from the associated textual description ($\vb{W} = \vb{w}_{0:n}$).
Given an event representation, an associated sequence (of words or images) can be obtained using a decoder $\Phi^{x-dec}$:
\begin{equation}
    \begin{split}
        \vb{x}_t | \vb{z}_t \sim \mathcal{N}(\vb{\mu}_{x,t}, \vb{\sigma}^2_{x,t})~\text{where}~[\vb{\mu}_{x,t}, \vb{\sigma}^2_{x,t}] = \Phi^{x-dec}(\vb{z}_t,\vb{x}_{t-1})
    \end{split}
\end{equation}
where $\vb{X} = \vb{x}_{0:T}$ may correspond to the flattened embedding of words $\vb{W}$ or images $\vb{V}$, and $\mathcal{N}(\cdot|\cdot)$ is a Gaussian distribution (assume prior) with parameters generated by the neural network $\Phi^{x-dec}$.
Thus, the resulting joint model mapped over trajectories $p(\vb{X}, \vb{z})$ factorizes as:
\begin{equation}
    p(\vb{x}_0)\prod_{t=1}^{T}p(\vb{x}_t|\vb{z}_{\leq t},\vb{x}_{<t})p(\vb{z}_t|\vb{z}_{<t})
    \label{eq:events-model}
\end{equation}
The functions $\Phi^{x-dec}$ and the transition function $p(\vb{z}_t^H|\vb{z}_{<t}^H)$ are approximated by sequence-to-sequence models (in this case transformers (\cite{vaswani2017attention})).
\newline\textbf{Encoding: }
An input sequence of video frames is downsampled to 200. The visual encoder $\vb{Z}_v~=~\vb{z}_{0:16}~\sim~q(\vb{z}_v|\vb{V})$ generates a sequence of event representations such that each event $\vb{z}_v\in \mathbb{R}^{768}$. 
Similarly, textual events $\vb{Z}_w=\vb{z}_{0:16}~\sim q(\vb{z}_w|\vb{W})$ are also generated using seq2seq transformer models.  
\newline\textbf{Decoding:}
We decode in a cross-modal manner, where the events abstracted from the visual domain are used to re-generate the textual description and \textit{vice-versa}. In what follows, prime notation refers to a re-generated value. Thus, the visual events are used to regenerate words using $\vb{W}'=\vb{w}'_{0:n}~\sim p(\vb{w}'|\vb{z}_v)$ and textual events are used to subsequently regenerate demonstration frame embedding $\vb{V}'=\vb{v}'_{0:m}~\sim p(\vb{v}'|\vb{z}_w)$.

\textbf{Learning Objective: }
We emphasize that we do not require supervision for temporal segmentation, i.e., we do not require annotations which demarcate the beginning and ending of a event, both in language and in the space of video frame timestamps.
Our approach uses several loss terms between network outputs to achieve our objective. The $\text{soft-DTW}$ (\cite{cuturi2017soft}) is used to compute the match between two sequences of varying length. It is calculated between several sequences to generate the pre-training loss term, $\mathcal{L}_{pretrain}$.
\begin{equation}
    \small
        \mathcal{L}_{pretrain}=\text{soft-DTW}(\vb{V},\vb{V}') + \text{soft-DTW}(\vb{Z}_v,\vb{Z}_w)+\text{soft-DTW}(\vb{W},\vb{W}')
    \label{eq:pre-training-objective}
\end{equation}
We posit that this loss function provides the inductive bias necessary for learning the event latent space. The term $\text{soft-DTW}(\vb{S},\vb{S}')$ ensures reconstruction of demonstration frames from the textual events, while $\text{soft-DTW}(\vb{W},\vb{W}'))$ ensures the generation of textual description from visual events. $\text{soft-DTW}(\vb{Z}_s,\vb{Z}_w)$ aligns the textual and visual event spaces.  
\subsection{Skill Learning using Cyclical Homomorphisms}
\begin{figure}[t]
    \centering
    \includegraphics[width = \textwidth, height=110pt]{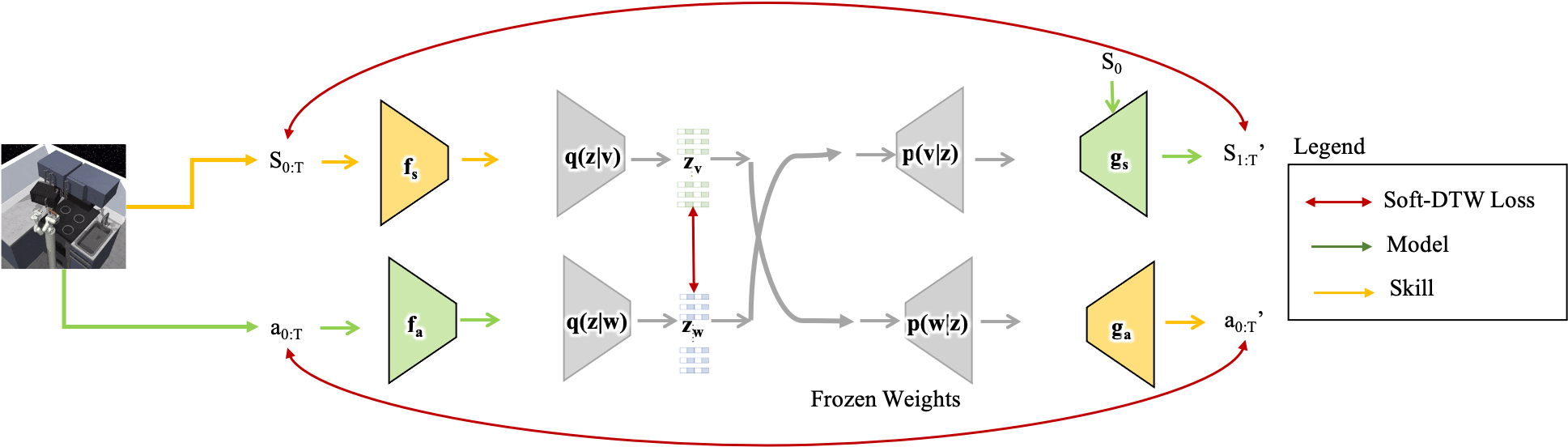}
    \caption{\textbf{Distillation.} We freeze the weights of the Backbone network and learn MDP homomorphisms from the MDP of the robotic kitchen domain to the abstract MDP of human demonstrations ($g$), and \textit{vice-versa} ($f$), in a self-supervised manner. The latent space simultaneously contains event representations from human cooking videos and skills from the robotic domain.}
    \label{fig:distillation}
\end{figure}
After pre-training on cooking videos of human demonstrations using Eq.~\ref{eq:pre-training-objective}, the weights of the encoders ($q(\vb{z}_v|\vb{v})$ and $q(\vb{z}_w|\vb{w})$) and decoders ($p(\vb{w}'|\vb{z}_{v})$ and $p(\vb{v}'|\vb{z}_{w})$) are frozen. Subsequently, offline demonstration (i.e., sequences of states and actions) in the robotic domain is used. This data consists of demonstrations by an expert robot performing tasks in the environment. In our case, the robot demonstration data consists of how to open a microwave oven, open cabinets, turn on the light, etc. Adapter functions $f~\text{and}~g$ are then learnt which map demonstration trajectories of states and actions from a trained robot performing these tasks in the environment onto the same space of video and word embeddings as used for human cooking.
\subsubsection{Skills and Environment Dynamics}
As in \cite{keele1968movement}, we define a skill as a sequence actions that may be executed in and of itself, without sensory feedback. Additionally, when a skill is applied to an environment, it results in a sequence of transitions that uniquely identifies the skill. For example, a skill which lifts an object in an environment is identified by both the sequence of actions applied by the agent and the resultant sequence of transitions in the environment. Thus, a latent vector $\vb{z}$ contextualizes both the policy ($\mathcal{\pi}_{\theta}$) and the subsequent model ($\vb{p}_{\phi}$):
\begin{equation}
        \vb{a}_t \sim \vb{\pi}_{\theta}(\vb{s}_{t}, \vb{z}_{t})~\text{and}~\vb{s}_{t+1} \sim \vb{p}_{\phi}(\vb{s}_{t}, \vb{z}_{t})
\end{equation}
Thus, the models over trajectories of states and actions in demonstration data factorize as:
\begin{equation}
    p(\vb{s}_0)\prod_{t=1}^{T}p(\vb{s}_{t+1}|\vb{z}_{t},\vb{s}_{t})p(\vb{z}_t|\vb{z}_{<t})~\text{and}~p(\vb{a}_0)\prod_{t=1}^{T-1}p(\vb{a}_{t+1}|\vb{z}_{t},\vb{a}_{\le t})p(\vb{z}_t|\vb{z}_{<t})
    \label{eq:dyanamics-model}
\end{equation}
\subsubsection{Cyclical Homomorphisms}
\begin{figure}[t]
    \centering
    \includegraphics[width = \textwidth, height=100pt]{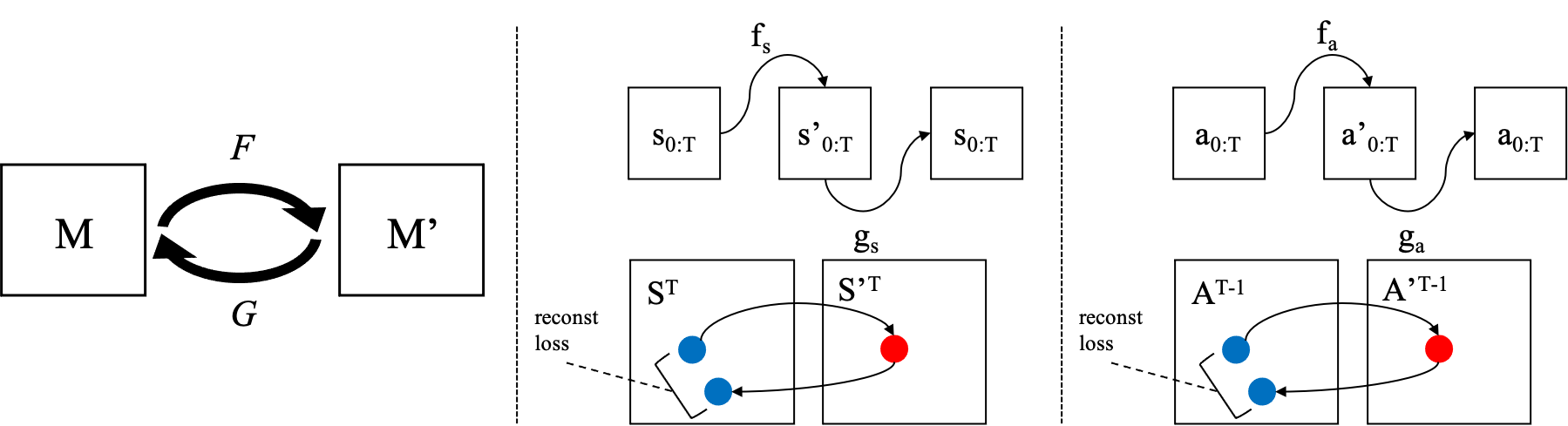}
    \caption{\textbf{Cyclical Homomorphisms.} During pre-training, \approachName learns an embedding space which encodes events occurring in an abstract MDP in which the human demonstrator behaves. This is done using video frames and textual commentary which serve as proxies for states and actions. Subsequently, a pair of homomorphisms map the robotic MDP into and out of the abstract MDP. These are learnt using a reconstruction loss.}
    \label{fig:cyclical_homomorphisms}
\end{figure}
Upon inspection, one can find that models in Eq.~\ref{eq:events-model} and in Eq.~\ref{eq:dyanamics-model} consist of the same structure, i.e., knowledge of the latent representation and of the history of a sequence determines the transition (with the exception of the state models in Eq.~\ref{eq:dyanamics-model} due to the Markov assumption, where rather than history, the current state is sufficient). Consider the pair of MDPs - $\mathbf{M}$ in the robotic domain ($\mathcal{S}$, $\mathcal{A}$) with state space $\mathcal{S}$, and action space $\mathcal{A}$ and second, $\mathbf{M}'$, the abstract MDP with state space $\mathcal{S}'=\omega(\mathcal{V})\times\xi(\mathcal{W})$ and action space $\mathcal{A}'=\Omega(\mathcal{V})\times\Xi(\mathcal{W})$ where $\mathcal{V}$ and $\mathcal{W}$ represent the spaces of video frames and words and $\Omega(.)$,$\omega(.)$,$\xi(.)$ and $\Xi(.)$ are unknown mappings from text and videos to state and  action representations (in what follows, knowing their exact form is not necessary). We exploit the shared structure between these MDPs to learn a pair of MDP homomorphisms - from the robotic MDP to the abstract MDP and \textit{vice-versa}. As defined in \cite{ravindran2004approximate}:
\begin{definition}[MDP Homomorphisms]
A Deterministic MDP homomorphism from an MDP $\mathcal{M}=(\mathcal{S},\mathcal{A},\mathcal{T},\mathcal{R})$ to an MDP $\mathcal{M'}=(\mathcal{S'},\mathcal{A'},\mathcal{T'},\mathcal{R'})$ is a tuple of functions, $\mathcal{F}=(\vb{f}_s, \vb{f}_a)$ with:
\begin{itemize}
    \item $\vb{f}_s:\mathcal{S} \to \mathcal{S}'$ the state embedding function, and
    \item $\vb{f}_a:\mathcal{A} \to \mathcal{A}'$ the action embedding function
\end{itemize}
such that the following identities hold:
\begin{equation}
    \forall \vb{s}_t,\vb{s}_{t+1}\in\mathcal{S},\vb{a}_t\in\mathcal{A}~\text{if}~\vb{s}_{t+1}= \mathcal{T}(\vb{s}_t,\vb{a}_t)~\text{then:}~\vb{f}_s(\vb{s}_{t+1})= \mathcal{T}'(\vb{f}_s(\vb{s}_{t}),\vb{f}_a(\vb{a}_{t}))
\end{equation}
\begin{equation}
    \forall \vb{s}_t\in\mathcal{S},\vb{a}_t\in\mathcal{A}, \mathcal{R}(\vb{s}_t,\vb{a}_t)= \mathcal{R}'(\vb{f}_s(\vb{s}_{t}),\vb{f}_a(\vb{a}_{t}))
\end{equation}
\end{definition}
Thus we learn a pair of MDP homomorphisms - $\mathcal{F}:\mathcal{M}\to\mathcal{M}'$ and $\mathcal{G}:\mathcal{M}'\to\mathcal{M}$ such that $d(\mathcal{G}(\mathcal{F}(\mathcal{M})), \mathcal{M})$ is minimized, where $d(\cdot, \cdot):\mathcal{M} \times \mathcal{M}\to \mathbb{R}_+$ is a suitable distance metric in the space of MDPs.
\subsubsection{Cyclical Homomorphic Objective}
We learn each of the homomorphisms by learning the state and action embedding functions separately. This is done by freezing the weights of the encoders ($q(\vb{z}_v|\vb{v})$ and $q(\vb{z}_w|\vb{w})$) and decoders ($p(\vb{w}'|\vb{z}_{v})$ and $p(\vb{v}'|\vb{z}_{w})$) in the backbone network pre-trained on the human demonstrations. Subsequently, $\vb{f}_s$, $\vb{f}_a$, $\vb{g}_s$, and $\vb{g}_a$ are learnt such that $\vb{f}_s$ and $\vb{f}_a$ (i.e., the forward homomorphism) map sequences of states and actions from demonstrations in the kitchen environment into the spaces of video frames and words respectively. Next the pre-trained encoders, i.e., ($q(\vb{z}_v|\vb{v})$ and $q(\vb{z}_w|\vb{w})$), generate the latent skill vectors for these trajectories. The skill vectors are fed into the pre-trained decoders ($p(\vb{w}'|\vb{z}_{v})$ and $p(\vb{v}'|\vb{z}_{w})$) to regenerate the sequences in the space of words and video frames. Finally, these sequences are fed back into the inverse MDP homomorphism ($\vb{g}_s$, and $\vb{g}_a$) to generate the original input sequence (Fig.~\ref{fig:cyclical_homomorphisms}).
\begin{equation}
    \small
        \mathcal{L}_{distil}=\text{soft-DTW}(\vb{S},\vb{S}') +\text{soft-DTW}(\vb{A},\vb{A}')
    \label{eq:pre-training-objective}
\end{equation}
\section{Experiments}
Our architecture is versatile, in that it results in the simultaneous learning of a latent conditioned dynamics model of the environment (green arrows in Fig.~\ref{fig:distillation}) and a latent conditioned policy or skill network (yellow arrows in Fig.~\ref{fig:distillation}).  
Through experiments, we present three main thrusts - \textbf{representation learning}, \textbf{dynamics learning} and \textbf{skill learning}. We study the utility of our approach in learning adaptable event representations. We show that our work results in unsupervised analogy learning of motion sequences. Subsequently, we study the ability of our latent conditioned model to utilize prior knowledge to quickly learn a long-horizon model of its environment, outperforming several state-of-the-art sophisticated baselines. Finally, we study the ability of \approachName to generate skills simply from human demonstration. We show that our agent performs complex motion behaviors in the robot kitchen environment akin to stirring, grasping, pouring, etc., which were demonstrated by humans in the cooking videos. 

We train our backbone network on the \textbf{YouCook2} (\cite{ZhXuCoCVPR18}) dataset which comprises instructional videos for 89 unique recipes ($\sim$22 videos per recipe) containing labels that separate the long horizon trajectories of demonstrations into events - with explicit time stamps for the beginning and end of each event along with the associated commentary. Subsequently, we train the cyclical homomorphic objective on demonstrations from \textbf{d4RL} dataset (\cite{fu2020d4rl}) of the \textbf{Franka Kitchen} environment (\cite{gupta2019relay}). The goal of the FrankaKitchen environment is to interact with the various objects to reach a desired state configuration. The objects the agent can interact with include the position of a kettle, flipping a light switch, opening and closing a microwave and cabinet doors, or sliding another cabinet door. The desired goal configuration for all 3 tasks is to complete 4 subtasks: open the microwave, move the kettle, flip the light switch, and slide open the cabinet door.   
\subsection{Long Horizon Dynamics Learning}
Dynamics learning, especially in an offline manner is a challenging endeavour. Dynamics learning using expressive models such as neural networks has proven to be challenging due to uncertainty stemming from insufficient data (epistemic uncertainty) and from the inherent stochasticity of an environment (aleatoric uncertainty). Further, long horizon dynamics modelling has long remained the bane of model-based reinforcement learning systems. Without an adequate model to rely upon while planning in the long horizon, model-based RL systems, although interpretable and simple have fallen behind recent advances in model-free RL. The root cause of many failures of long-horizon planning is error aggregation, i.e., sub-optimal predictions at each time step during inference results in a trajectory of states that moves increasingly further from the ground truth transitions in an environment with time. 

Here, instead of inferring a long-horizon trajectory of states in an auto-regressive manner by passing actions into the model one-by-one, we propose to feed a whole sequence of actions into \approachName. \approachName decodes a sequence of skill representations from such a sequence of actions and subsequently generates the expected trajectory of resultant states conditioned on a starting state. 

We compare our approach to four popular model/dynamics learning approaches currently used as state-of-the-art. As defined in \cite{chua2018deep}: 
    \newline \textbf{Probabilistic Neural Network (PNN):} A probabilistic NN is a network whose output
    neurons simply parameterize a probability distribution function, capturing aleatoric uncertainty. We use the negative log prediction probability as our loss function, i.e., $loss_{PNN}~=~\-\sum_{i=1}^Nlog(\vb{f}_{\theta}(\vb{s}_t+1|\vb{s}_t, \vb{a}_t))$ and choose the output distribution to be Gaussian with a diagonal covariance matrix.
    \newline \textbf{Determinstic Neural Network (DNN):} A deterministic NN is a special case of a probabilistic network that outputs delta distributions centered around point predictions denoted by $\vb{f}_{\theta}(\vb{s}_t, \vb{a}_t)=\vb{f}_{\theta}(\vb{s}_{t+1}|\vb{s}_t, \vb{a}_t)=Pr(\vb{s}_{t+1}|\vb{s}_t, \vb{a}_t)=\vb{\delta}(\vb{s}_{t+1}-\vb{f}_{\theta}(\vb{s}_t,\vb{a}_t))$. It is trained using $loss_{DNN}~=~\-\sum_{i=1}^N||\vb{s}_{t+1}-\vb{f}_{\theta}(\vb{s}_t, \vb{a}_t))||_{2}$. MSE can be interpreted as $loss_{PNN}$ with a Gaussian model of fixed unit variance, but cannot be used in practice for propagation. 
    \newline \textbf{Ensembles - PE and DE: }As in \cite{chua2018deep}, we consider ensembles of $B$-many bootstrap models, using $\theta_{B}$ to refer to the parameters of our $b^{th}$ model $\vb{f}_{\theta_{b}}$. Ensembles can consist of probabilistic NNs or deterministic NN both with effective probabilty distributions as $\vb{f}_{\theta} = \frac{1}{b}\sum_1^b\vb{f}_{\theta_b}$.

\approachName is pre-trained on the YouCook2 \cite{Zhou_2018_CVPR} demonstrations. Subsequently, we provide each of the baselines and \approachName with a dataset of demonstrations in the robot kitchen environment. Each of the baselines and \approachName are trained for 1, 5, and 10 epochs and subsequently evaluated on unseen data for a 2-step, 5-step and 180-step (full sequence) next-state prediction. We repeat training over 10 random seeds and report standard error across the seeds. 

We find that our model is robust to long-horizon error aggregation. In Table \ref{Dynamics-Learn}, we compare the ability of our model to quickly adapt to dynamics of the kitchen environment when the backbone network is pre-trained on cooking videos. We find that our model outperforms all current state-of-the-art approaches in learning dynamics of the environment faster and maintaining performance over longer horizons.     
\begin{table}[t]
  \caption{\textbf{Long Horizon Dynamics Learning.} We study the ability of \approachName to learn long horizon models of an environment in an offline manner. We pre-train the Backbone networks and on the cyclical homomorphic objective for $1, 5$ \text{and} $10$ epochs. We find that \approachName performs up to 10 times better over long-horizon sequences \textbf{(lower RMSE is better)}.}
  \label{Dynamics-Learn}
  \centering
  \begin{tabular}{llll}
    \toprule
    Method     & 2-Step     & 5-Step & Full Sequence\\
    \midrule
    \textbf{PNN: }\cite{higuera2018synthesizing}&&&\\
    \quad 1 Epoch & $1.0065 \pm 0.2719$  & $0.6447 \pm 0.0886$ &$0.9131 \pm 0.0126$   \\
    \quad 5 Epoch & $0.7297 \pm 0.1462$  & $0.9212 \pm 0.25$ &$0.8612 \pm 0.0132$   \\
    \quad 10 Epoch & $0.7658 \pm 0.2858$  & $1.2529 \pm 0.3738$ &$0.6937 \pm 0.0131$   \\
    \midrule
    \textbf{DNN: }\cite{nagabandi2018neural}&&&\\
    \quad 1 Epoch & $0.4184 \pm 0.0009$  & $0.4189 \pm 0.0009$ &$0.4502 \pm 0.0013$   \\
    \quad 5 Epoch & $0.3897 \pm 0.0027$  & $0.407 \pm 0.0016$ &$0.4334 \pm 0.0017$   \\
    \quad 10 Epoch & $0.1327 \pm 0.0076$  & $0.2496 \pm 0.0113$ &$0.3357 \pm 0.0081$   \\
    \midrule
    \textbf{DE: }\cite{lakshminarayanan2016simple}&&&\\
    \quad 1 Epoch & $0.4201 \pm 0.0011$  & $0.4205 \pm 0.001$ &$0.451 \pm 0.0011$   \\
    \quad 5 Epoch & $0.3912 \pm 0.0018$  & $0.4078 \pm 0.0013$ &$0.4343 \pm 0.0014$   \\
    \quad 10 Epoch & $0.1384 \pm 0.0036$  & $0.257 \pm 0.0054$ &$0.341 \pm 0.0043$   \\
    \midrule
    \textbf{PE: }\cite{chua2018deep}&&&\\
    \quad 1 Epoch & $0.5339 \pm 0.0376$  & $0.5061 \pm 0.02$ &$0.57 \pm 0.0041$   \\
    \quad 5 Epoch & $0.5124 \pm 0.0524$  & $0.5633 \pm 0.06$ &$0.5434 \pm 0.005$   \\
    \quad 10 Epoch & $0.3585 \pm 0.0757$  & $0.3925 \pm 0.0581$ &$0.4327 \pm 0.0026$   \\
    \midrule
    \textbf{\approachName (ours)}\\
    \quad 1 Epoch & $0.2872 \pm 0.0069$  & $0.2946 \pm 0.006$ &$0.3077 \pm 0.0012$   \\
    \quad 5 Epoch & $0.0818 \pm 0.003$  & $0.0675 \pm 0.0023$ &$0.0655 \pm 0.0001$   \\
    \quad 10 Epoch & $0.0552 \pm 0.003$  & $0.052 \pm 0.0037$ &$0.0511 \pm 0.0001$   \\
    \bottomrule
  \end{tabular}
\end{table}

\subsection{Unsupervised Analogy Learning}
\approachName is trained on two separate datasets - \textbf{YouCook2} and \textbf{d4RL-FrankaKitchen}. During pre-training, the agent learns environment-agnostic event representations encoding the associated sequences of video frames and textual commentary from \textbf{YouCook2}. During homomorphism learning, it learns to map sequences of states and actions in the robotic environment into the same latent space using the \textbf{d4RL-FrankaKitchen} dataset. Thus, we obtain a \textbf{shared latent space}, which contains event representations from cooking videos and kitchen environment skill representations. In Fig.~\ref{fig:analogy_learning}, we plot a reduced dimensional t-SNE plot (\cite{van2008visualizing}). We then explore the overlapping latent vectors in the plot and decode them to visualize the analogies discovered by the architecture. We find that \approachName models motion programs successfully across domains without any supervision. The model learns to pick up on analogies between a spreading motion in the cooking videos to a horizontal sliding motion in the kitchen demonstration sequences. In other instances, it discovers analogies between circular hinge motions in the kitchen environment and circular stirring motions in the cooking videos. We emphasize, no supervision was provided to map the individual domains to one another. The cyclical pair of MDP homomorphisms resulted in an unsupervised analogy discovery. 
\begin{figure}[t]
    \centering
    \includegraphics[width = \textwidth, height= 230pt]{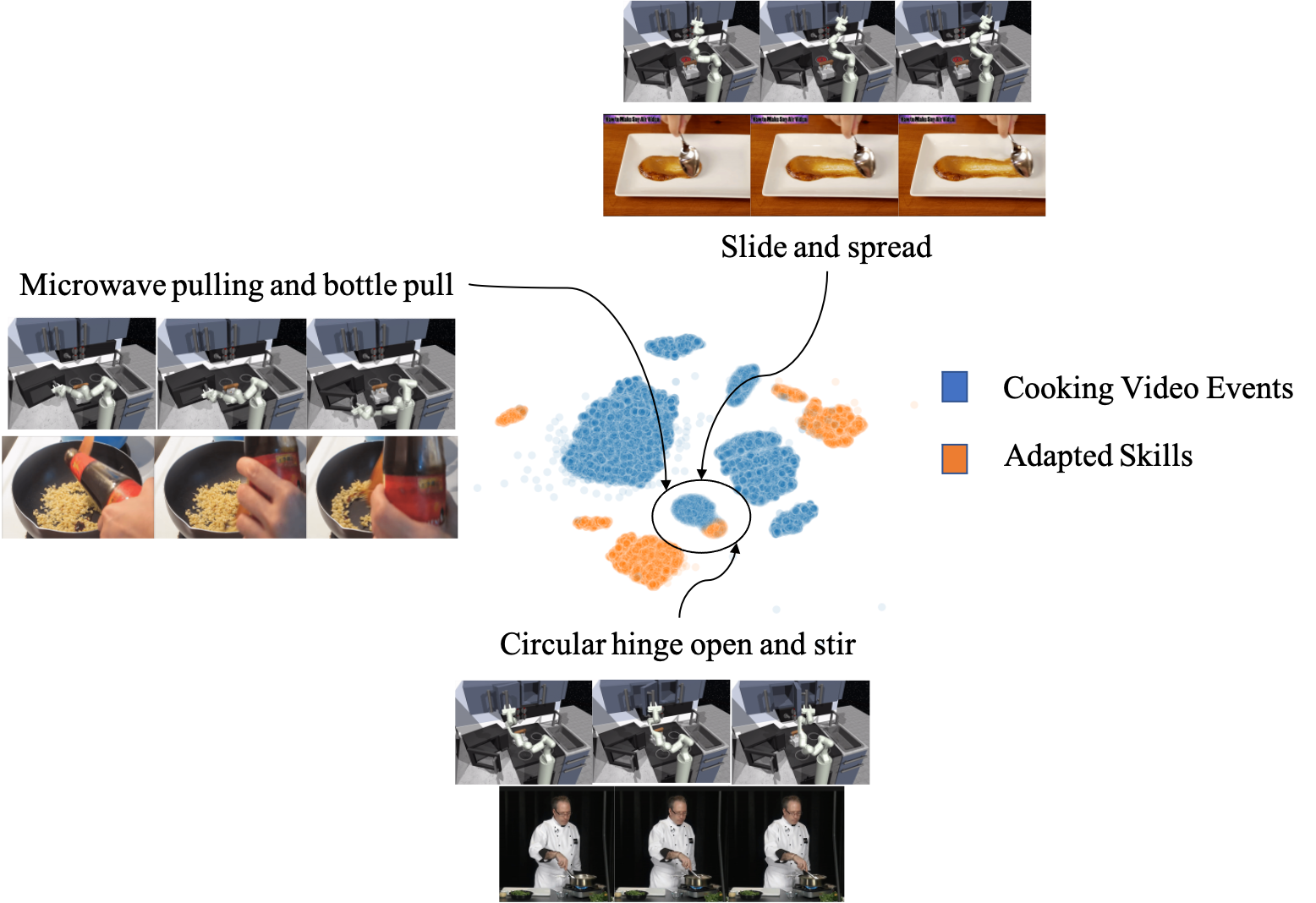}
    \caption{\textbf{Unsupervised Analogy Learning.} Using the cyclical homomorphisms, we embed events from the human cooking demonstration videos and skills from the robotic kitchen environment into the same latent space. We explore the representation learning capacity by finding overlapping regions of the latent space and exploring their semantic meaning. \approachName produces semantically meaningful analogies.}
    \label{fig:analogy_learning}
\end{figure}

\subsection{Zero-shot Skill Generation}
\label{sec:zeroshotskill}
\begin{figure}[t]
    \centering
    \includegraphics[width = \textwidth, height=150pt]{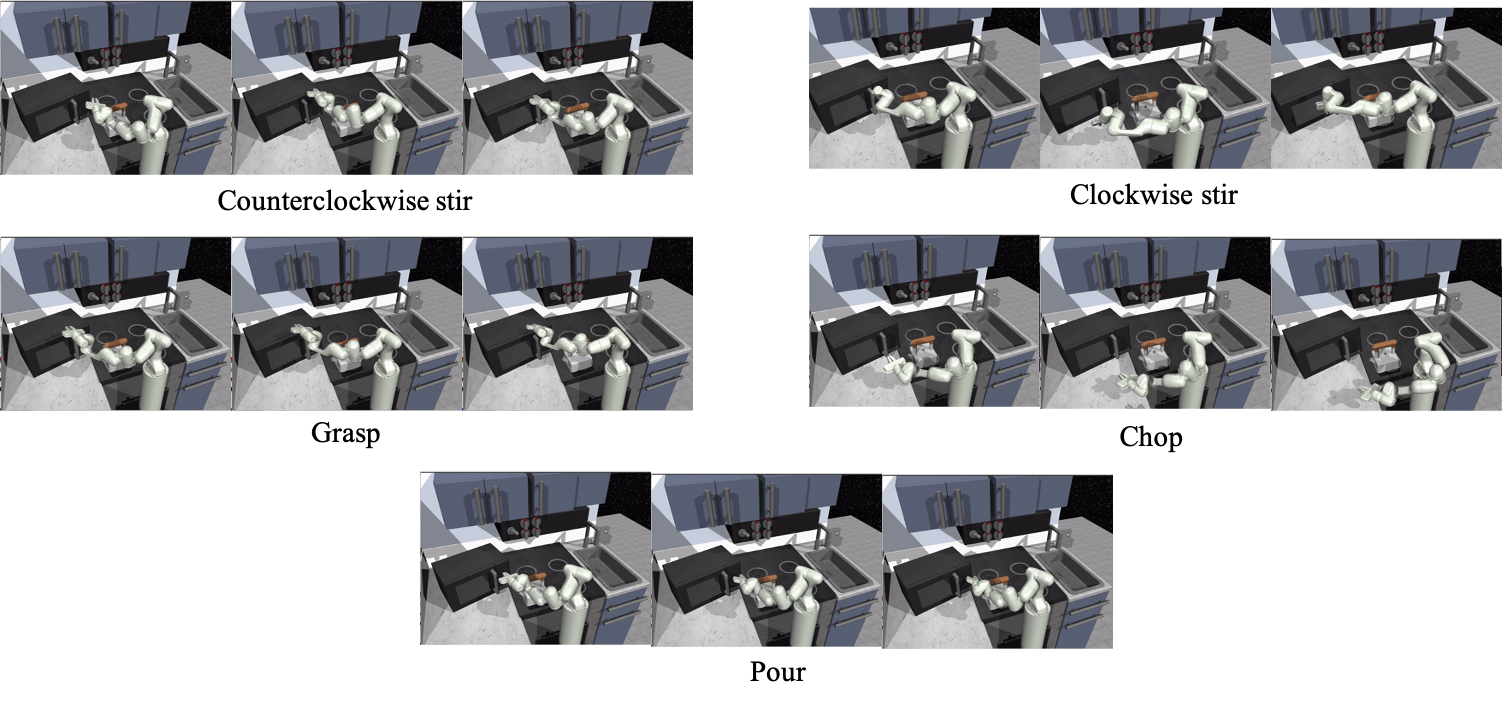}
    \caption{\textbf{Qualitative Evaluation of Generated Skills.} \approachName generates several significant semantically meaningful skills merely from human demonstrations. These motions are learnt in a reward free manner and can be used for complex tasks. Click \href{https://drive.google.com/drive/folders/1glA6_KjMGcXI0q10Z-kA51LLOkrNncyE?usp=sharing}{\textbf{here}} to view gifs of discovered skills.}
    \label{fig:analogy_learning}
\end{figure}
\begin{figure}[t]
    \centering
    \includegraphics[scale=0.4]{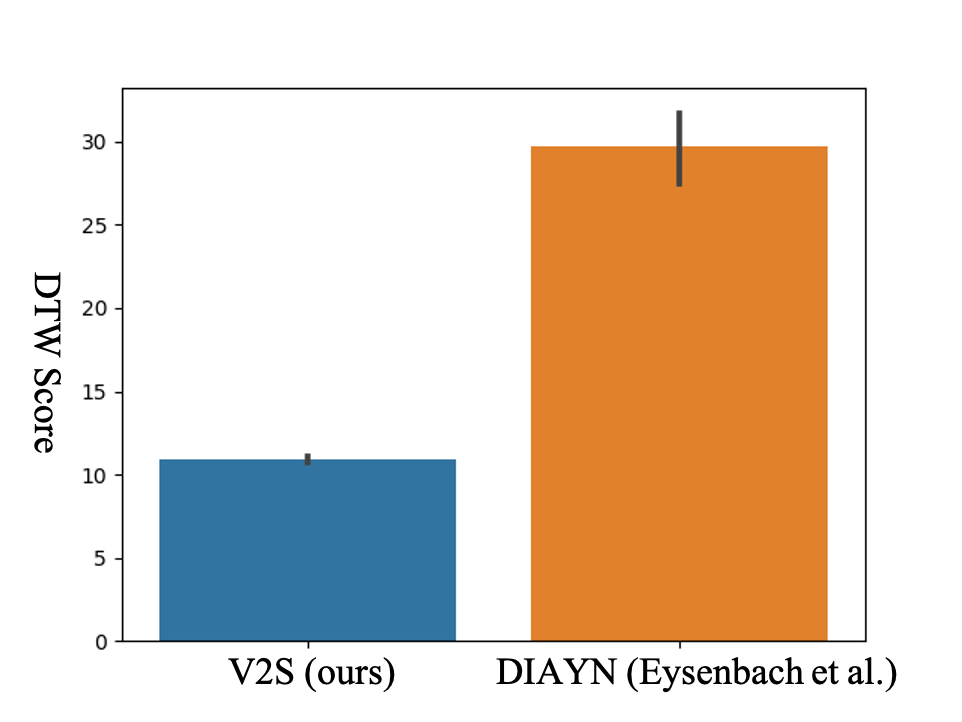}
    \caption{\textbf{Quantitative Evaluation of Generated Skills.} We study the time-warped sequence distance \textbf{(lower is better)} between demonstrations from an expert agent in the kitchen environment and the skills generated by \approachName. We find that \approachName generates skills closer to expert trajectories than \cite{eysenbach2018diversity}, a state-of-the-art unsupervised skill learning approach.}
    \label{fig:skill_quality}
\end{figure}
As both video demonstration event representations and skill vectors are embedded in a \textbf{shared latent space}, we explore the ability of our architecture to generate useful and semantically meaningful skills in a zero-shot manner. To do this, we sample event representation vectors and pass them as input to the textual decoders $p(\vb{w}'|\vb{z}_{v})$ and subsequently, to the action embedding of the Inverse Homomorphism $\mathcal{G}'=$, $\vb{g}_a$. The resultant actions are then applied to the environment to visualize how knowledge acquired from the cooking videos can be used to learn long horizon action sequences that are semantically meaningful. 

We find that the model is able to generate complex skills that were never seen in the robotic demonstration data, but were demonstrated by humans in the cooking video data. For example, our model produces a robotic stirring motion both clockwise and counter-clockwise. Other skills include motion sequences that could be used for grasping, pouring, etc. if the robot was given extra artifacts like cups, water, etc.  \href{https://drive.google.com/drive/folders/1glA6_KjMGcXI0q10Z-kA51LLOkrNncyE?usp=sharing}{\textbf{This link}} shows discovered skills from human demonstrations. 

\subsection{Quantitative Skill Assesment}
In Fig.~\ref{fig:skill_quality}, we study the utility of the skills generated by \approachName in being able to effectively manipulate objects and successfully complete tasks in the environment. To this end, we decode each of the generated skills from our architecture in terms of control signals and find their smallest sequence discrepancy in the demonstration data. This discrepancy calculation is performed using the Dynamic Time-Warping loss proposed by \cite{cuturi2017soft}. This allows us to calculate the sequence matches between 2 sequences of varying lengths. We compare the quality of our skills to those generated by DIAYN (\cite{eysenbach2018diversity}). We find that our skills are up to three times closer to demonstration trajectories than those generated by DIAYN. We repeat experiments over 6 random seeds.  
\section{Conclusion}
We propose a reward-free approach to skill learning, which utilizes prior knowledge to aid decision-making in a complex environment. We show that our architecture results in powerful long-horizon models and semantically meaningful skills and uses human demonstration data to aid both. A drawback of our architecture is the size and training time (several GPU-months of training means significant energy expenditure); work towards leaner models will be beneficial. Additionally, there is still a gap between demonstration and generated skills (Fig.~\ref{fig:skill_quality}). Work towards bridging this gap is necessary.   
\begin{ack}
Use unnumbered first level headings for the acknowledgments. All acknowledgments
go at the end of the paper before the list of references. Moreover, you are required to declare
funding (financial activities supporting the submitted work) and competing interests (related financial activities outside the submitted work).
More information about this disclosure can be found at: \url{https://neurips.cc/Conferences/2021/PaperInformation/FundingDisclosure}.

Do {\bf not} include this section in the anonymized submission, only in the final paper. You can use the \texttt{ack} environment provided in the style file to autmoatically hide this section in the anonymized submission.
\end{ack}

\bibliographystyle{plainnat}
\bibliography{sample}





\section*{Checklist}

The checklist follows the references.  Please
read the checklist guidelines carefully for information on how to answer these
questions.  For each question, change the default \answerTODO{} to \answerYes{},
\answerNo{}, or \answerNA{}.  You are strongly encouraged to include a {\bf
justification to your answer}, either by referencing the appropriate section of
your paper or providing a brief inline description.  For example:
\begin{itemize}
  \item Did you include the license to the code and datasets? \answerYes{See Section~\ref{gen_inst}.}
  \item Did you include the license to the code and datasets? \answerNo{The code and the data are proprietary.}
  \item Did you include the license to the code and datasets? \answerNA{}
\end{itemize}
Please do not modify the questions and only use the provided macros for your
answers.  Note that the Checklist section does not count towards the page
limit.  In your paper, please delete this instructions block and only keep the
Checklist section heading above along with the questions/answers below.

\begin{enumerate}

\item For all authors...
\begin{enumerate}
  \item Do the main claims made in the abstract and introduction accurately reflect the paper's contributions and scope?
    \answerYes
  \item Did you describe the limitations of your work?
    \answerYes{See Conclusion.}
  \item Did you discuss any potential negative societal impacts of your work?
    \answerYes{See conclusion.}
  \item Have you read the ethics review guidelines and ensured that your paper conforms to them?
    \answerYes
\end{enumerate}

\item If you are including theoretical results...
\begin{enumerate}
  \item Did you state the full set of assumptions of all theoretical results?
    \answerNA{}
	\item Did you include complete proofs of all theoretical results?
    \answerNA{}
\end{enumerate}

\item If you ran experiments...
\begin{enumerate}
  \item Did you include the code, data, and instructions needed to reproduce the main experimental results (either in the supplemental material or as a URL)?
    \answerYes{Code link in supplementary material}
  \item Did you specify all the training details (e.g., data splits, hyperparameters, how they were chosen)?
    \answerYes{}
	\item Did you report error bars (e.g., with respect to the random seed after running experiments multiple times)?
    \answerYes{}
	\item Did you include the total amount of compute and the type of resources used (e.g., type of GPUs, internal cluster, or cloud provider)?
    \answerYes{}
\end{enumerate}

\item If you are using existing assets (e.g., code, data, models) or curating/releasing new assets...
\begin{enumerate}
  \item If your work uses existing assets, did you cite the creators?
    \answerYes{}
  \item Did you mention the license of the assets?
    \answerYes{In code link in Supplementary Material.}
  \item Did you include any new assets either in the supplemental material or as a URL?
    \answerYes{Supplementary Material has code. Sec~\ref{sec:zeroshotskill} has link to Gifs. }
  \item Did you discuss whether and how consent was obtained from people whose data you're using/curating?
    \answerNA{}
  \item Did you discuss whether the data you are using/curating contains personally identifiable information or offensive content?
    \answerNA{}
\end{enumerate}

\item If you used crowdsourcing or conducted research with human subjects...
\begin{enumerate}
  \item Did you include the full text of instructions given to participants and screenshots, if applicable?
    \answerNA{}{}
  \item Did you describe any potential participant risks, with links to Institutional Review Board (IRB) approvals, if applicable?
    \answerNA{}
  \item Did you include the estimated hourly wage paid to participants and the total amount spent on participant compensation?
    \answerNA{}
\end{enumerate}

\end{enumerate}


\appendix
\newpage
\section{Implementation Details}
We down-sample video frames per trajectory to 200 frames and encode each frame with ResNet-32 (pretrained on MSCOCO dataset) (\cite{he2016deep}) to a $512\times{}1$ dimension embedding. Comments are encoded using BERT-base pre-trained embeddings with a $768$ hidden dimension. Each of the $p(\vb{z}|\vb{w})$, $p(\vb{z}|\vb{s})$, $p(\vb{s}|\vb{z})$, $q(\vb{w}|\vb{z})$ modules consist of the Transformer (\cite{vaswani2017attention}) Encoder with 8 hidden layers and 8-Head Attention which takes as input, a positionally-encoded sequence and outputs attention weights. It is then passed through a Transformer Decoder with 8 hidden layers to generate latent variables having dimension $event length\times{768}$.
The event length is 16 events per trajectory. 

We keep the maximum number of events discovered to 16.
These assumptions are based on the YouCook2 dataset statistics where the minimum number of segments were 5 and the maximum as 16.
We train the network with Adam optimizer for 100 epochs with $lr=1e-5$, $\alpha = 1$ and $\beta = 1$ for all our experiments along with a batch-size of 128. We use 16x Nvidia A100 GPUs to train the backbone network. 
For the robotic data, we sample sequences of 180 states and 179 action as input. The batch size is fixed to 32. The training takes $~5$ days for training the backbone and an additional $~3$ days to train the various adapted versions.  

\end{document}